\documentclass[sigconf]{acmart}

\AtBeginDocument{%
  }
    
\usepackage{multirow}
\usepackage{algorithm}
\usepackage{algpseudocode}

\usepackage{xcolor}

\newtheorem{theorem}{Theorem}

\newcommand{\modelname}{MR-Traj}
\newcommand{\modelnamespace}{MR-Traj }

\copyrightyear{2026}
\acmYear{2026}
\setcopyright{cc}
\setcctype{by}
\acmConference[KDD 2026] {Proceedings of the 32nd ACM SIGKDD Conference on Knowledge Discovery and Data Mining V.2}{August 9--13, 2026}{Jeju Island, Republic of Korea.}
\acmBooktitle{Proceedings of the 32nd ACM SIGKDD Conference on Knowledge Discovery and Data Mining V.2 (KDD 2026), August 9--13, 2026, Jeju Island, Republic of Korea}
\acmISBN{979-8-4007-2259-2/2026/08}
\acmDOI{10.1145/3770855.3818889}

\begin{document}

\title{Coarse-to-Fine Multi-Resolution Diffusion Models for  Trajectory Generation in Urban Systems}




\author{Wen Ye}
\authornote{Both authors contributed equally to this research.}
\email{yewen@usc.edu}
\orcid{0009-0006-6196-0824}
\affiliation{%
  \institution{University of Southern California}
  \city{Los Angeles}
  \state{California}
  \country{USA}
}

\author{Muyan Weng}
\authornotemark[1]
\email{muyanwen@usc.edu}
\orcid{0009-0005-8723-0689}
\affiliation{%
  \institution{University of Southern California}
  \city{Los Angeles}
  \state{California}
  \country{USA}
}

\author{Chuizheng Meng}
\email{chuizhem@usc.edu}
\orcid{}
\affiliation{%
  \institution{University of Southern California}
  \city{Los Angeles}
  \state{California}
  \country{USA}
}

\author{Hao Niu}
\email{ha-niu@kddi.com}
\orcid{}
\affiliation{%
  \institution{KDDI Research, Inc.}
  \city{Saitama}
  \country{Japan}
}

\author{Yizhou Zhang}
\email{zhangyiz@usc.edu}
\orcid{}
\affiliation{%
  \institution{University of Southern California}
  \city{Los Angeles}
  \state{California}
  \country{USA}
}

\author{Yan Liu}
\email{yanliu.cs@usc.edu}
\orcid{0000-0002-7055-9518}
\affiliation{%
  \institution{University of Southern California}
  \city{Los Angeles}
  \state{California}
  \country{USA}
}

\renewcommand{\shortauthors}{Trovato et al.}

\begin{abstract}
Understanding human mobility is critical for a wide range of urban applications, including traffic management, epidemic control, and urban planning. However, due to privacy concerns, the availability of large-scale public trajectory data remains limited, posing challenges for downstream mobility analysis. Existing methods for synthetic trajectory generation primarily focus on matching global distribution similarity, while often overlooking mobility patterns across different spatial and temporal resolutions that are essential for practical utility.

To address these challenges, we propose a novel multi-resolution diffusion framework, MR-Traj, for large-scale trajectory generation. MR-Traj explicitly models trajectories as compositions of coarse-grained milestones and fine-grained segments, enabling the capture of complex spatial-temporal dependencies at multiple resolutions. Experimental results demonstrate that MR-Traj achieves comparable performance to state-of-the-art methods in terms of global distribution similarity, while consistently outperforming them in modeling fine-resolution mobility patterns and supporting downstream urban mobility tasks. In addition, by introducing stochasticity at multiple resolution levels, MR-Traj generates more diverse trajectories, which empirically reduces trajectory linkage risk under a seed-guided data release setting. Our code is available at \url{https://github.com/Ray0202/MR-Traj}.
\end{abstract}



\maketitle
\section{Introduction}

Human mobility data provides critical insights into population-level movement patterns and supports a wide range of applications, including urban planning \cite{xu2019anomalous,9561461}, traffic estimation \cite{li2020trajectory,griesemer2024active}, epidemic control \cite{kraemer2020effect}, and location-based services \cite{zhang2022counterfactual}. In urban and transportation systems, such data plays a central role in city-scale decision making and mobility analysis. However, access to large-scale mobility data is often restricted in practice due to privacy and regulatory constraints, particularly for organizations that collect extensive trajectory records but cannot publicly release them without risking sensitive information disclosure. As a result, synthetic trajectory generation has emerged as a practical alternative, enabling data sharing while avoiding the direct exposure of individual-level records.

Early approaches to synthetic trajectory generation primarily relied on Markov Chain models. With advances in deep learning, generative models such as Variational Autoencoders (VAEs) \cite{doersch2016tutorial} and Generative Adversarial Networks (GANs) \cite{creswell2018generative} were subsequently explored, often combined with recurrent architectures such as Long Short-Term Memory networks (LSTMs) to capture temporal dependencies \cite{sherstinsky2020fundamentals,zhang2023dp,rao2020lstm}. Despite their success, many existing methods depend on strong prior assumptions or handcrafted semantic features, such as home/work locations or points of interest \cite{long2023practical,feng2020learning}, and frequently struggle to reproduce fine-grained mobility patterns at high spatial resolution \cite{wang2021large,niu2022mu2rest}. This limitation is particularly important in urban-scale settings, where both city-wide structure and neighborhood-level movement details are necessary for reliable analysis.

More recently, diffusion models have emerged as a promising alternative, offering stable training dynamics and flexible conditioning mechanisms for controlled trajectory generation \cite{zhu2023difftraj,ye2024domain}. Nevertheless, diffusion-based trajectory generation is typically guided by metadata from original trajectories, which may still introduce privacy risks when synthetic data is released. At the same time, accurately modeling urban mobility requires capturing patterns across multiple spatial and temporal resolutions, where global movement patterns describe long-range structure while local patterns reflect fine-grained behaviors within specific regions.

Another limitation of existing methods lies in their predominant focus on minimizing global distribution discrepancies between real and synthetic trajectories, with relatively limited consideration of downstream analytical utility and responsible data release \cite{choi2021trajgail,kang2020trag,feng2020learning,cao2022dslob}. In practice, many mobility analysis tasks operate on partial trajectories or localized segments, making fidelity at finer resolutions equally important and motivating the need for multi-resolution modeling in which global structure and local movement are jointly represented. Hierarchical models that decompose complex processes into multiple probabilistically connected levels provide a principled framework for describing stochastic spatio-temporal phenomena \cite{arab2008hierarchical,cao2021spectral}. 

In parallel, recent work has explored large-scale pre-trained trajectory models that learn transferable mobility representations from massive unlabeled data, enabling improved generalization across regions and supporting downstream tasks such as prediction, completion, and generation \cite{wu2024pretrained,zhu2024unitraj,garg2025gps,cao2025conversational,cao2023tempo}.

Building on these observations, recent studies have shown that integrating diffusion models into multi-resolution or coarse-to-fine frameworks can effectively capture structured dependencies in complex data \cite{qiang2023coarse,pi2023hierarchical}. For example, HGHOI generates high-level milestones before modeling detailed motions in human-object interaction \cite{pi2023hierarchical}, while HierDiff adopts a hierarchical diffusion strategy for molecular generation \cite{qiang2023coarse}. In the context of urban mobility, separately modeling abstract structural information (e.g., milestones) and detailed trajectory segments provides a principled way to capture multi-scale movement patterns while increasing diversity in the generated data.

Motivated by these insights, we propose a \textbf{M}ulti-\textbf{R}esolution framework for modeling complex spatio-temporal dependencies in mobility \textbf{Traj}ectories using diffusion models (\modelname). We focus on a two-level resolution setting, where a milestone-level diffusion model captures global trajectory structure and a segment-level diffusion model generates fine-grained movements conditioned on synthesized milestones. This formulation enables \modelnamespace to preserve both large-scale mobility structure and local movement details, while reducing the risk of overly deterministic trajectory reproduction under metadata-guided generation.

In summary, our contributions are threefold:
\begin{itemize}
\item We propose a multi-resolution diffusion framework for synthetic urban trajectory generation, jointly modeling global milestone trajectories and local trajectory segments under metadata guidance.
\item We conduct extensive experiments on large-scale real-world mobility datasets, showing that our method achieves performance comparable to state-of-the-art approaches in global distribution similarity while consistently improving local distribution fidelity.
\item We demonstrate that \modelnamespace supports downstream urban mobility analysis tasks, including taxi destination prediction and anomalous mobility pattern identification, while exhibiting lower empirical linkage risk than existing metadata-guided baselines.
\end{itemize}

\section{Related Work}

Early work on synthetic trajectory generation primarily modeled trajectories as sequences of locations governed by Markovian assumptions, where each location depends only on the previous one \cite{song2003evaluating, meng2025physics}. Extensions such as Input-Output Hidden Markov Models relax uniform transition assumptions by incorporating contextual information into latent processes \cite{yin2017generative}. Despite their interpretability, Markov-based approaches rely on autoregressive sampling, which limits scalability and prevents efficient parallel generation of long trajectories.

With the rise of deep learning, generative models such as Variational Autoencoders (VAEs) \cite{doersch2016tutorial} and Generative Adversarial Networks (GANs) \cite{creswell2018generative} have been widely explored for trajectory synthesis. Recurrent architectures, including Long Short-Term Memory networks (LSTMs), are often combined with GANs to capture temporal dependencies \cite{sherstinsky2020fundamentals,rao2020lstm,zhang2023dp}. Representative methods include TrajGANs \cite{liu2018trajgans}, which first demonstrated GAN-based trajectory generation, and subsequent extensions that incorporate urban structure priors or multi-stage generation pipelines to improve realism at different spatial resolutions \cite{feng2020learning,wang2021large}. Reinforcement learning formulations, such as SeqGAN, further model trajectory generation as a sequential decision process \cite{yu2017seqgan}. However, GAN-based models are often difficult to train, computationally expensive at scale, and offer limited controllability over generated trajectories, making them less suitable for large-scale urban mobility data.

VAEs have also been adapted to trajectory generation by introducing hierarchical latent structures. For example, VOLUNTEER employs a two-layer VAE that first infers user-level latent attributes and then generates trajectories conditioned on these attributes \cite{long2023practical}. While VAEs provide a principled probabilistic framework, they often produce overly smooth trajectories and rely on restrictive distributional assumptions, which can limit their ability to capture complex and fine-grained mobility patterns.

Recently, diffusion models have emerged as a promising alternative for trajectory generation, offering stable training dynamics and flexible conditioning mechanisms~\cite{cao2025timedit,cao2026pinfdit}. Diffusion-based approaches have demonstrated strong performance in modeling complex, high-dimensional structured data, including spatio-temporal trajectories \cite{yang2024survey}. In particular, DiffTraj shows that diffusion probabilistic models can generate synthetic trajectories that closely match real mobility distributions under metadata guidance \cite{zhu2023difftraj}.

Beyond single-resolution generation, a growing body of work explores hierarchical and multi-resolution modeling strategies to capture the multi-scale nature of mobility data. Coarse-to-fine diffusion frameworks decompose trajectory generation into global planning and local refinement stages, enabling better modeling of long-range structure and local movement details \cite{guo2025leveraging}. Related approaches combine diffusion with structured planning or multi-scale representations to model trajectories alongside other spatio-temporal signals, such as traffic flows \cite{luo2025traveller,liu2025multi}. Recent studies further investigate controllable diffusion for structured or categorical trajectory data and emphasize evaluation through downstream urban analysis tasks, rather than relying solely on distribution-level similarity \cite{dirmeier2024synthetic,song2024controllable,kapp2023generative,graser2025mobilitydl}.

Our work builds on these advances by explicitly incorporating the hierarchical structure of spatio-temporal mobility data into a multi-resolution diffusion framework. By modeling global trajectory structure and local trajectory segments separately, our approach aims to better capture multi-scale dependencies inherent in real-world urban mobility, improving both generation fidelity and downstream analytical utility.

\section{Preliminaries}
\subsection{Problem Definition}
\label{sec:problem}

A Trajectory that is sampled at regular time intervals is defined as $\mathcal{S} = [s_1,...,s_n] $ where $s_i = (lat_i,lon_i)$ represents the location of the vehicle at timestamp $i$ when $s$ is on a continuous spatial scale; $s_i = gid_i$ represents the grid id that the taxi is in at timestamp $i$ on a discretized spatial scale; $n$ denotes the total number of points in the trajectory. This dual representation caters to different needs in spatial data processing, with latitude and longitude providing precise locations and grid ids facilitating spatial analysis on a coarser scale and evaluation concerned with distribution similarity.

The objective of Synthetic Trajectory Generation is to generate synthetic trajectories $S' = [s_1',...s_n']$ where $S'$ preserves the spatial-temporal characteristics and patterns observed in $S$. These synthetic trajectories should be highly useful for downstream spatial-temporal analysis, meaning they should mimic real-world mobility patterns and can be used to support tasks such as traffic management, and mobility behavior analysis. In general, many downstream tasks of synthetic trajectory generation involves predicting some properties of a trip, denoted as $g(S_u, S_o)$ such that the downstream model uses only a segment of the full trajectory where \( S_o \) denote the observed portion of a trajectory, while \( S_u \) denote the unobserved portion to be inferred or completed. 

\subsection{Theoretical Motivation}
\begin{theorem}
If the target function \( g(S_u, S_o) \) is \( L \)-Lipschitz and \( D \) represents the diameter of the support set of \( S_u \), the empirical risk \( R \) of a predictor trained on synthetic data \( P'(S) \) and evaluated on real data \( P(S) \) is bounded as:
\[
\begin{aligned}
    R &\leq L D^2 \| P(S_u, S_o) - P'(S_u, S_o) \|_{TV} \\
    &+ L D^2 \| P'(S_o) - P(S_o) \|_{TV} + C
\end{aligned}
\]

where: \( \| \cdot \|_{TV} \) denotes the Total Variation (TV) distance, and \( C \) is a constant.
\end{theorem}
 The upper bound derived in theorem 1 indicates that minimizing the empirical risk of a downstream model requires reducing the TV distances between both the joint and marginal distributions of the synthetic and real data. While global similarity between the real and synthetic distributions $L D^2 \| P(S_u, S_o) - P'(S_u, S_o) \|_{TV}$ is often emphasized in existing methods, our theoretical results suggest that minimizing this distance, though essential, is not sufficient for fully optimizing downstream performance. Existing methods which focus solely on reducing the global distance between \( P \) and \( P' \) may not maximize utility. The theoretical result indicates that utility is influenced not only by global distributional similarity but also by local trajectory patterns where the observed fraction of trajectories could be any segment of the complete trajectory. This motivates us to employ a multi-resolution framework to model different resolutions of a trajectory and hence modeling both complete trajectories and different resolutions of the trajectory such as milestones and segments. 
 The details of this proof can be found in the supplementary material where we define the optimal predictor given the objective and empirical risk; leverage the Lipschitz Condition \cite{boyd2004convex} and bound the Wasserstein Distance \cite{villani2009optimal} between two distributions; apply triangle inequality on the Total Variance Distance \cite{devroye2013probabilistic} bound. 
 
 We further motivate our design choice from the privacy consideration perspective. Modeling trajectories in a multi-resolution fashion allows us to obtain increased diversity of generated trajectories through multi-level noise injection. By introducing noise at various resolution of the synthetic trajectory generation process, the model effectively creates a broader range of possible synthetic outcomes. 

\section{Methodology}
\begin{figure*}
    \centering
    \includegraphics[width=\linewidth]{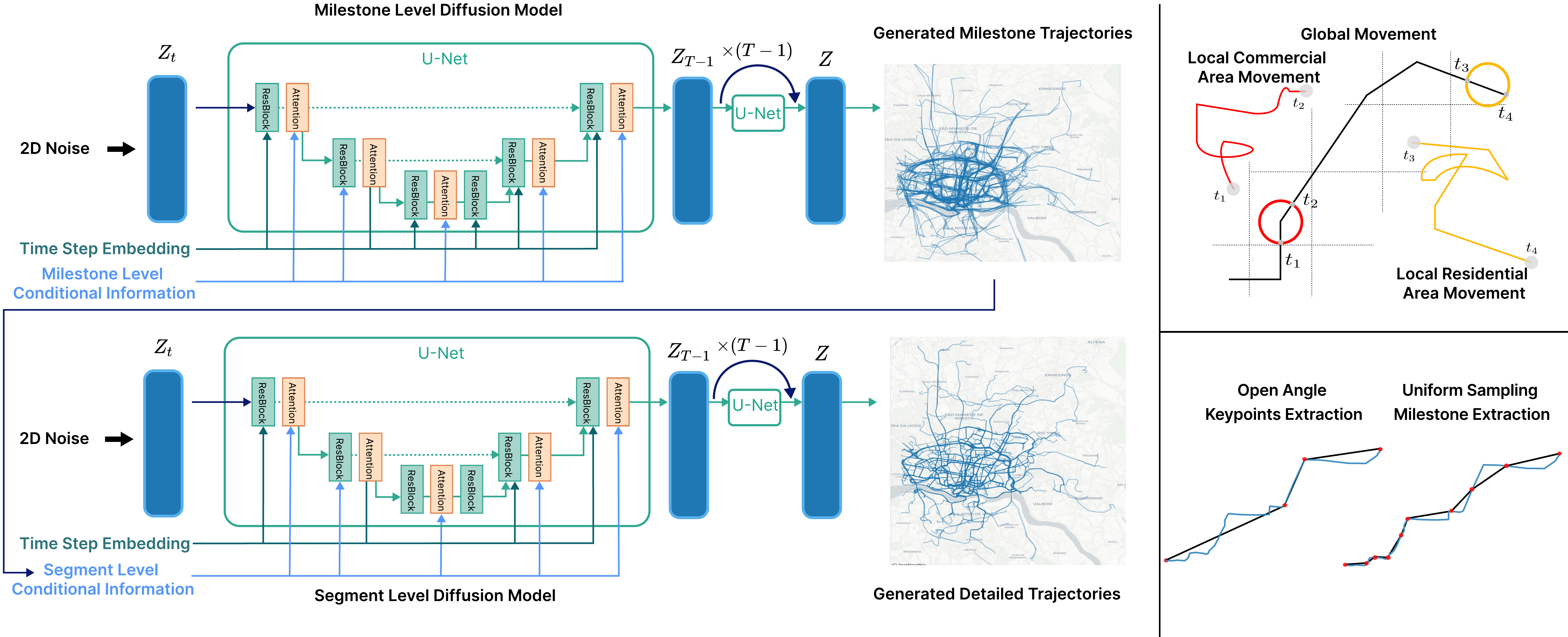}
    \caption{Left: Multi-Resolution Framework of Two-level Diffusion Model; Right top: Example of Global Movement and Local Movement Patterns; Right bottom: Illustration of two different Milestone Extraction strategies}
    \label{fig:model}
\end{figure*}
In this section, we formally introduce the multi-resolution framework inspired by the motivating analysis. 
We employ a coarse-to-fine strategy to generate synthetic trajectories. The multi-resolution framework is composed of two models: a milestone-level diffusion model and a segment-level diffusion model. The classic Origin-Destination Matrix can be seen as the most abstract level of resolution of a trajectory; milestones then is at a finer resolution and segments provide the finest resolution. Both models leverage the U-Net architecture as their backbone but are conditioned on different levels of trajectory resolution to focus on distinct aspects of trajectory generation. Figure \ref{fig:model} shows the overall structure of the multi-resolution diffusion model framework. The milestone-level model is designed to learn the broader trajectory distribution, capturing high level distributions of trajectories, i.e. general direction of moving. Conversely, the segment-level model is tasked with detailing the finer aspects of the trajectory, filling in the gaps between milestones with realistic local movement patterns.

Our choice of backbone for both diffusion models is the widely acclaimed U-Net structure \cite{ronneberger2015u}. We adopt this structure along with a condition embedding module implemented as a straightforward feedforward neural network from prior work \cite{zhu2023difftraj}. The condition module embeds trajectory attributes like departure time, distance traveled, start and end locations, enabling the diffusion model to learn the conditional distribution of trajectories based on these attributes. We further modify the noise schedule of diffusion model to adhere to the cosine schedule rather than the linear schedule. This choice is motivated by the cosine schedule's efficiency, allowing for a more gradual addition of noise and thus spending fewer timesteps on purely noisy transformations \cite{nichol2021improved}. 



\subsection{Milestone Extraction \& Learning Milestone Trajectories}

We define a milestone trajectory as an abstraction of the complete trajectory at both temporal and spatial resolutions. In general, an Origin-Destination pair represents the lowest-resolution milestone, while the complete trajectory corresponds to the highest resolution. Milestone derivation is a key component of the framework, and we consider two strategies: uniform sampling and open angle extraction. Open angle extraction, such as the two-pass corner detection algorithm, is commonly used in trajectory compression tasks \cite{sun2016overview} to identify keypoints based on open angles formed by neighboring points and local curvature. A visual illustration of the two milestone extraction techniques is shown in Figure~\ref{fig:model}. 

We derive milestones from original trajectories using a simple yet effective uniform sampling strategy, selecting a point every $\alpha$ points as a milestone. This design targets dense GPS trajectories without explicit semantic or POI annotations (e.g., taxi mobility data), where milestones act as geometric anchors rather than semantic activity boundaries.
We denote the extracted milestone trajectory as
\begin{equation}
    \mathcal{M} = [m_1, ... , m_{\lceil n//\alpha \rceil}] = [s_1, s_{\alpha}, ..., s_{\lceil n//\alpha \rceil \alpha}]
\end{equation}
While keypoint-based extraction provides semantically meaningful compression by capturing directional changes, we find uniform sampling more effective for our setting, as demonstrated in the ablation study in Section~5. Uniform sampling preserves the overall trajectory shape and standardizes time intervals between milestones, which simplifies segment-level diffusion modeling. By enforcing equal-length segments, we reduce the complexity of modeling variable-length trajectories and enable more consistent and effective synthetic trajectory generation, ultimately improving data quality and downstream utility.

Overall, the denoising process of the milestone-level diffusion model is defined as
\begin{equation}
    p_{\theta}(\mathcal{M}_{t-1}|\mathcal{M}_{t},c_m) = \mathcal{N}(\mathcal{M}_{t-1}; \mu_{\theta}(\mathcal{M}_t, t|c_m), \Sigma_{\theta}(\mathcal{M}_t, t|c_m))
\end{equation}
where $\mathcal{M}_t$ denotes a noisy milestone trajectory at diffusion timestep $t$, and $c_m$ represents milestone-level (global) conditions constructed by concatenating attributes such as travel time, travel distance, start location, and end location. The training objective of the milestone-level diffusion model is
\begin{equation}
    \mathcal{L}(\theta) = \mathbb{E}_{t, \mathcal{M}_0, \epsilon} \left[ \| \epsilon - \epsilon_{\theta}(\mathcal{M}_t, t|c_m) \|^2 \right]
\end{equation}
where $\epsilon \sim \mathcal{N}(0, I)$ is the Gaussian noise injected in the forward diffusion process, and $\epsilon_{\theta}(\cdot)$ denotes the noise prediction network parameterized by $\theta$.

\subsection{Learning Milestone Conditioned Segment}
The milestone extraction operation on a single trajectory also results in $\lceil n//\alpha\rceil-1$ segments of the original trajectory, we denote the segments as:
\begin{equation}
    \mathcal{S} = [ \mathcal{S}_1, ... , , \mathcal{S}_{\lceil n//\alpha \rceil-1} ]
\end{equation}
\begin{equation}
\begin{aligned}
    \mathcal{S}_i &= [s_{i\alpha}, s_{i\alpha+1}, ..., s_{(i+1)\alpha-1}, s_{(i+1)\alpha}] \\
    &= [m_i,s_{i\alpha+1}, ..., s_{(i+1)\alpha-1}, m_{i+1}]
    \end{aligned}
\end{equation}
The generation of trajectory segments between milestones is conditioned on segment-level conditions as well as its corresponding milestones as start and end guidance (every segment starts and ends with a milestone point). This process ensures that the segments are not only realistic in isolation but also coherent when considered as part of the overall trajectory. The denoising process of the segment-level diffusion model can be described as
\begin{equation}
    p_{\theta}(\mathcal{S}_{j,t-1}|\mathcal{S}_{j,t},c_{s,j}) = \mathcal{N}(\mathcal{S}_{j,t-1}; \mu_{\theta}, \Sigma_{\theta})
\end{equation}
\begin{equation}
    \mu_{\theta}\leftarrow\mu_{\theta}(\mathcal{S}_{j,t}, t|c_{s,j}), \Sigma_{\theta} \leftarrow \Sigma_{\theta}(\mathcal{S}_{j,t}, t|c_{s,j})
\end{equation}
\begin{equation}
    c_{s,j} = [c_j, m_j, m_{j+1}] 
\end{equation}
where $\mathcal{S}_{j,t}$ denotes the noisy $j$-th segment trajectory at diffusion timestep $t$, $c_{s,j}$  denotes the concatenated segment level conditions consisting of attribute information $c_j$ calculated from $\mathcal{S}_j$ and $m_j,m_{j+1}$ which denotes the milestones used for guidance. The subscript $j$ specifies the segment index within a complete trajectory and $m_j,m_{j+1}$ is the corresponding generated milestones that specify the start and end location of the $j$-th segment trajectory. During training, the milestones $m_j$ are extracted from original trajectories. The training objective of the segment level diffusion model becomes
\begin{equation}
    \mathcal{L}(\theta) = \mathbb{E}_{t, \mathcal{S}_0, \epsilon} \left[ \| \epsilon - \epsilon_{\theta}(\mathcal{S}_{j,t}, t|c_{s,j}) \|^2 \right]
\end{equation}
\subsection{Synthetic Trajectory Generation}

During inference, we use the same attribute information as in training.
However, the milestones guiding segment generation are replaced with synthetic milestones produced by a milestone-level diffusion model.
Details of the algorithm and variable-length trajectory handling are provided in Algorithm~\ref{alg:MYALG}.
Variable-length trajectories are handled via interpolation-based resampling during both training and inference, consistent with Algorithm~\ref{alg:MYALG}. 

We adopt an efficient sampling scheme from prior work to avoid expensive sampling at every diffusion step~\cite{song2020denoising}.
Specifically, for each sampled noise, we denoise every $s$ timesteps during the backward process with stride $s$.
As our framework conditions on multiple attributes, the generated synthetic trajectories can be viewed as seed trajectories~\cite{bindschaedler2016synthesizing}, where each corresponds to a real-world trajectory.
The combination of all given conditions defines a semantic class.
Synthetic trajectories are generated under these conditions and can therefore replace original trajectories within the same semantic class, yielding a synthetic dataset that preserves the semantic information of the original dataset~\cite{bindschaedler2016synthesizing}. 

\begin{algorithm}[htbp]
\caption{\modelnamespace algorithm}
\textbf{Input} Trajectory $T$, Uniform Sampling Rate $\alpha$, Attribute Information $C$ \\
\textbf{Output} Synthetic Trajectory $T'$
\begin{algorithmic}[1]
    \State $T$ is an original trajectory of length $n$
    \State $\mathcal{M} = [m_1,\dots,m_{n//\alpha}]$ is the extracted milestone trajectory
    \State $\mathcal{S} = [\mathcal{S}_1, \dots, \mathcal{S}_{n//\alpha-1}]$ is the set of segments between consecutive milestones
    \State $length(\mathcal{S}_i) = \alpha + 1$
    \State Initialize $C_m = C =$ [total distance, total time, \dots, start location, end location]
    \State Initialize $C_s = [\ ]$
    \For{$\mathcal{S}_i$ in $\mathcal{S}$}
        \State compute attributes for $\mathcal{S}_i$ based on $C$ and append to $C_s$
    \EndFor 
    \State $\mathcal{M}$ is resampled to the $0.75$-th quantile length of all milestone trajectories in the dataset via interpolation
    \State Train MDM($\mathcal{M}, C_m$) \Comment{Milestone-Level Diffusion Model} 
    \State Train SDM($\mathcal{S}_i, C_s[i]$) \Comment{Segment-Level Diffusion Model} 
    \State $\mathcal{M}' =$ sample from MDM($z, C_m$) \Comment{$z$ is random noise}
    \State $\mathcal{M}'$ is resampled back to length $n//\alpha$ by interpolation
    \For{$i$ in length of $\mathcal{S}$}
        \State $C_s[i][[-2,-1]] = [\mathcal{M}'[i], \mathcal{M}'[i+1]]$ 
        \Comment{replace start and end location guidance for segment $i$}
        \State $\mathcal{S}_i' =$ sample from SDM($z, C_s[i]$) \Comment{$z$ is newly sampled noise}
    \EndFor
    \State $T'$ is obtained by concatenating all segments in $\mathcal{S}' = [\mathcal{S}'_1,\dots,\mathcal{S}'_{n//\alpha -1}]$
\end{algorithmic}
\end{algorithm}
\label{alg:MYALG}

\section{Experiment}
In this section, we evaluate our model in comparison to various baseline model on two taxi trajectories dataset: Chengdu \cite{zhu2024synmob} and Porto \cite{pkdd-15-predict-taxi-service-trajectory-i}. Specifically, we  evaluate the similarity in distribution between the original and synthetic trajectories and also the downstream utility of generated synthetic trajectories. We further perform privacy assessment for seed synthetic trajectories generated under the guidance from original metadata. We considered first-order Markov Model: Markov \cite{song2003evaluating}, Hidden Markov Model: IOHMM \cite{yin2017generative}, LSTM Model: LSTM \cite{hochreiter1997long}, GAN based Model: Movesim \cite{feng2020learning}, Diffusion based Model: DiffTraj \cite{zhu2023difftraj}.
\begin{table*}[htbp]
  \caption{Distribution Evaluation of Generated Synthetic Trajectories compared to Original Trajectories. (\textbf{Bold}: the best result. \underline{Underline}: the 2nd best result.)'.' notation stands for Error}
  \label{tab:distribution}
  \scalebox{0.85}{
  \centering
  \begin{tabular}{ccccccccccc}
    \toprule
    \multirow{2}{*}{Dataset} & \multirow{2}{*}{Model} & \multicolumn{4}{c}{Global}& \multicolumn{4}{c}{Local}\\ \cline{3-10}
    & & Density. ($\downarrow$) & Distance. ($\downarrow$)& Trip. ($\downarrow$) & Pattern Score ($\uparrow$)& Density. ($\downarrow$)& Distance. ($\downarrow$) & Trip. ($\downarrow$) & Pattern Score ($\uparrow$)\\ 
    \midrule
     \multirow{6}{*}{\rotatebox[origin=c]{90}{Porto}} &\modelname& \underline{0.0015}& \textbf{0.0002}&\underline{0.0079}& \underline{0.91}&\textbf{0.0257} & \textbf{0.0039} &\textbf{0.1060}&\textbf{0.72}\\
     &DiffTraj&\textbf{0.0010} & \underline{0.0004} & \textbf{0.0061} & \textbf{0.93}&\underline{0.0371}&\underline{0.0044}&\underline{0.1480}&\underline{0.59} \\
     &Markov & 0.0175& 0.5250& 0.1505& 0.59&0.1730&0.6090 &0.6560&0.26\\
     &LSTM &0.0103 & 0.0257 & 0.0746 & 0.70& 0.1948&0.0643&0.6124&0.25\\
     &MoveSim & 0.2816 & 0.0463& 0.2977& 0.02&0.6635 &0.3210&0.6811 &0.03 \\
     &IOHMM & 0.0412& 0.0215 & 0.2867 & 0.34 &0.2060&0.0401 &0.5858&0.23\\
    \midrule 
    \multirow{6}{*}{\rotatebox[origin=c]{90}{Chengdu}} &\modelname & \underline{0.0028}& \underline{0.0035}&\textbf{0.0134}&\underline{0.81}& \textbf{0.0081}& \underline{0.0096}& \textbf{0.0395}& \textbf{0.75}\\
     &DiffTraj& \textbf{0.0018} & \textbf{0.003}  & \underline{0.0246}  & \textbf{0.94}& \underline{0.0182}& \textbf{0.0009}& \underline{0.0474}&\underline{0.66}\\
     &Markov & 0.0356 & 0.0373 & 0.5835& 0.27& {0.0772}&0.6930 & 0.5649 &{0.32}\\
     &LSTM & 0.0658 & 0.335 & 0.266 & 0.14& 0.0978 & {0.5504}&{ 0.5864}& 0.33\\
     &MoveSim &  0.0585 & 0.156 & 0.3852 & 0.13 & 0.1407  &0.0506& 0.4084 & 0.22\\
     &IOHMM  & 0.0757  & 0.0725  & 0.46  & 0.11& 0.2482& 0.6931& 0.5916&0.02\\
     
     \bottomrule
  \end{tabular}}
\end{table*}
\subsection{Distribution Similarity Evaluation}

The two large-scale taxi trajectory GPS dataset we used is the Chengdu Dataset and Porto Dataset.  We filtered out very short trajectories less than 10 points and focused on trajectories in the center region of the city of Porto, Portugal, also referred to as kernel region of a city \cite{kang2020trag}. This region is characterized by a higher and more uniform density of trajectories, offering a representative sample for analysis. The Chengdu dataset is already confined to a high density region and no preprocessing is performed.

To quantify the performance of \modelnamespace and compare it with baseline models, we employed several metrics detailed as the following. For metrics that ask for a discretization of the spatial space, we used a 16 $\times$ 16 grid over the spatial scale, resulting in grid resolution less than $1km^2$ per cell. The discretization is done only for evaluation purpose, the generated trajectories are made up of precise GPS locations. For detailed explanation on the evaluation metrics, please refer to the supplementary material. In addition to the global space, we also sampled trajectory segments within two popular regions from each city: Ribeira region in the city of porto and Taikoo Li commercial region in the city of Chengdu. We chose the two regions because they are highly popular regions within the two city with either tourist attractions and commercial venues. We perform the same evaluation on the segments within the chosen region. For baselines that do not generate segments, we use uniform sampling with the same sampling rate to define milestones which in turn defines segments from a trajectory. For detailed information on bounding boxes of each region and each dataset as well as additional meta information, please refer to the supplementary material.

Table \ref{tab:distribution} shows the evaluation result of \modelnamespace and other baselines. We achieve close parity with the best-performing baseline while outperforming all other baselines on global distribution similarity. The evaluation indicates that \modelnamespace closely matches the performance of the best-performing baseline DiffTraj model across all metrics on the global scale. However, when evaluated on regional segment similarity, \modelnamespace demonstrated superior performance on both datasets and all metrics. This result validated that \modelnamespace is detail oriented and successfully captures local details.

\subsection{Utility and Privacy Evaluation}


\begin{table}[t]
  \caption{
  Downstream utility evaluation of generated synthetic trajectories.
  For Taxi Destination Prediction, HDE denotes the Haversine Distance Error.
  For Anomalous Mobility Pattern Similarity, GH and WH denote the cosine similarity
  between anomalous mobility patterns at the global-hourly and weekly-hourly scales, respectively. P in metric names indicates Porto and C indicates Chengdu.
  }
  \label{tab:downstream}
  \centering
  \setlength{\tabcolsep}{4pt}
  \small
  \begin{tabular}{l|cc|ccc}
    \toprule
    \multirow{2}{*}{Model} 
    & \multicolumn{2}{c|}{Destination Prediction} 
    & \multicolumn{3}{c}{Anomalous Pattern Similarity} \\
    & P-HDE$\downarrow$ & C-HDE$\downarrow$ 
    & P-GH$\uparrow$ & P-WH$\uparrow$ & C-GH$\uparrow$ \\
    \midrule
    Original & 2.319 & 2.810 & 1.000 & 1.000 & 1.000 \\
    \midrule
    \modelname & \textbf{2.361} & \textbf{2.833} & \textbf{0.846} & \textbf{0.700} & \textbf{0.747} \\
    DiffTraj  & 2.371 & 2.896 & 0.835 & 0.678 & 0.659 \\
    Markov    & 2.389 & 2.970 & 0.698 & 0.463 & 0.397 \\
    LSTM      & 2.383 & 3.195 & 0.731 & 0.481 & 0.319 \\
    MoveSim   & 2.447 & 3.082 & 0.419 & 0.330 & 0.202 \\
    IOHMM     & 2.541 & 2.990 & 0.778 & 0.633 & 0.238 \\
    \bottomrule
  \end{tabular}
  \vspace{-15pt}
\end{table}

 In this section, we scrutinize the practical utility of the synthetic data produced by our model, \modelname, through its application to two critical downstream tasks: Taxi Destination Prediction and Anomalous Mobility Pattern Identification. These tasks were selected based on their relevance to the real-world applicability of synthetic trajectory data. The former represents a direct application of taxi trajectory data, crucial for traffic management and location-based services. The latter tests the synthetic data's ability to preserve the original dataset's unique or anomalous patterns, a vital aspect of data utility in detecting and analyzing mobility behaviors. Though normal mobility patterns are important, it's relatively easy to learn since it constitutes the majority of the dataset. But abnormal patterns can easily be neglected while showing some meaningful insights into atypical behaviors or events. To the best of our knowledge, the ability of a model to preserve original anomalous patterns in synthetic trajectories is highly underexplored. 
 
 Another highly considered aspect of synthetic trajectories is privacy and we further examine this crucial aspect by training a classifier on original data to link trajectory to users. By introducing noise at two levels, \modelnamespace demonstrates more diversity in the generated trajectories which becomes highly beneficial when it comes to privacy preservation without compromising the utility of generated trajectories.

\subsubsection{Taxi Destination Prediction}

Given a partial taxi trajectory $S_p = [s_1,...,s_j], j<n, S_p\subset S$ where $S$ is defined in section \ref{sec:problem}, we want to predict $s_n$, the destination point of the trajectory. The task involves predicting the end point of a taxi trip given a partial trajectory. This challenge is not only about understanding the spatial components of the trip but also about inferring the likely destination based on observed patterns in the data. We adopt T-CONV\cite{lv2018t} which is a convolutional neural network built for taxi trajectory prediction. Please refer to the supplementary material for details of T-CONV model. The T-CONV model is trained separately on the original and each generated synthetic datasets. The evaluation metric is the mean Haversine distance between the real destination and the predicted location for trajectories in the test set from original data.

Table \ref{tab:downstream} presents the Haversine distance errors for models trained on both original and synthetic data. \modelnamespace demonstrates a remarkable capability in maintaining high quality to the original data's spatial-temporal patterns, as evidenced by its Haversine Distance Error of 2.361. This metric, only slightly higher than the 2.319 error observed with the original data, suggests that the synthetic trajectories generated by \modelnamespace closely resemble the real-world movements, capturing essential dynamics necessary for accurate destination prediction. The minimal discrepancy between the haversine distance error between \modelnamespace and original data indicates that \modelnamespace effectively encapsulates the variability and complexity of urban mobility patterns in its synthetic outputs. Notably, \modelnamespace outperforms all baselines, further validating the utility of \modelnamespace generated data. 

\subsubsection{Anomalous Mobility Pattern Identification}
Urban mobility flows typically exhibit certain motifs or regular patterns reflecting the routine movements of people within an environment \cite{schneider2013unravelling}.  However, alongside these regular patterns, there are anomalous mobility patterns that deviate from the norm, often emerging during holidays, special events, extreme weather, or other unique circumstances \cite{xu2019anomalous}. Recognizing these anomalies is essential for urban planning, traffic management, and emergency response, allowing for preemptive action in anticipation of similar future events. Following prior work, we implemented a method to calculate mobility vectors and identify anomalous mobility patterns \cite{xu2019anomalous}. For detailed description of the procedure, please refer to the supplementary material.The number of mobility vectors depend on the temporal scale of interest. In this experiment setting, we examined global hourly scale where we calculate a mobility for each hour; and a weekly hourly scale where a mobility vector is calculated for each hour within each week. Our analysis focuses evaluating how effectively the synthetic trajectories generated by \modelnamespace retain these anomalous patterns present in the original dataset compared to other baselines. This evaluation is conducted using the pairwise cosine similarity between anomalous mobility vectors derived from both synthetic and original data.



\begin{figure}[htbp]
    \centering
    \includegraphics[width=\linewidth]{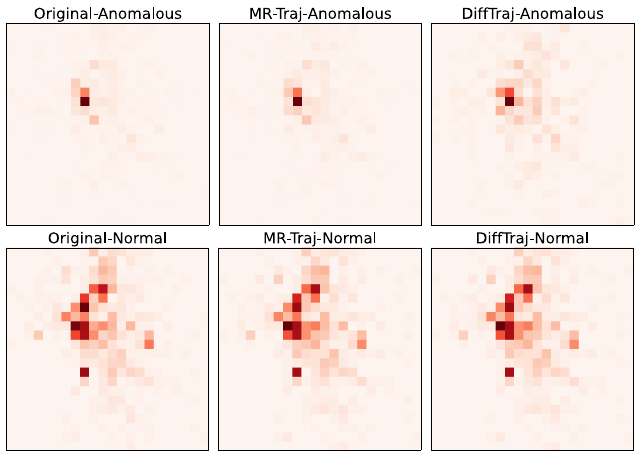}
    \caption{Comparison of normal and abnormal mobility patterns on discretized grid of Porto region from 1) Original trajectories, 2) \modelnamespace generated trajectories, and 3) DiffTraj generated trajectories}
    \label{fig:normal_abnormal}
    \vspace{-10pt}
\end{figure}

Table \ref{tab:downstream} presents the cosine similarity scores for anomalous mobility vectors derived from synthetic and original trajectories, calculated on both global hourly and weekly hourly scales. The Chengdu dataset does not have the Weekly Hourly Scale evaluation because there is no departure date information in the dataset. So, the seed trajectories cannot be grouped by week. \modelnamespace achieves the highest cosine similarity at both temporal scales. These results illustrate \modelname's proficiency in capturing and preserving anomalous mobility patterns present in the original trajectory data, with its performance notably surpassing that of other models. Figures \ref{fig:normal_abnormal} visually compare mobility vectors from both normal and anomalous clusters, derived from original, \modelname, as well as the best performing baseline DiffTraj generated synthetic data. These visual comparisons further highlight the model's ability to replicate both regular and anomalous urban mobility patterns effectively. 

\begin{table}[t]
\caption{Trajectory User Linking evaluation on seed synthetic trajectories. A lower metric value indicates a higher level of privacy. Other baselines are emitted in this evaluation because their synthetic generation process is not guided by specific metadata from original trajectory. The standard deviation is included in parenthesis as a result of 3 different runs with different numpy random seeds.}
\centering
\scalebox{0.8}{
\begin{tabular}{c|ccc}
\toprule
\textbf{Metric} & \textbf{Original Data} & \textbf{DiffTraj} & \textbf{\modelname} \\
\midrule
Acc ($\downarrow$) & 0.9738 (0.0031) & 0.8018 (0.0050) & \textbf{0.7325 (0.0069)} \\
Acc@5 ($\downarrow$) & 0.9986 (0.0003) & 0.9092 (0.0004) & \textbf{0.8668 (0.0013)} \\
F1 Score ($\downarrow$) & 0.9793 (0.0031) & 0.8142 (0.0043) & \textbf{0.7556 (0.0038)} \\
Precision ($\downarrow$) & 0.9781 (0.0035) & 0.8226 (0.0060) & \textbf{0.7686 (0.0042)} \\
Recall ($\downarrow$) & 0.9816 (0.0021) & 0.8152 (0.0024) & \textbf{0.7533 (0.0034)} \\
\bottomrule

\end{tabular}}
\label{tab:privacy}
\vspace{-10pt}
\end{table}

\subsubsection{Privacy Evaluation}Privacy is a crucial aspect of data management, especially in trajectory data, where personal and sensitive information can be inferred from movement patterns. Existing methods often overlook privacy concerns by focusing solely on distribution similarity and utility. These methods tend to ensure that synthetic data maintains statistical resemblance to the original data while maximizing usability, but fail to adequately safeguard individual privacy. 

In our privacy evaluation, we employ the MARC (Multiple-Aspect tRajectory Classifier) algorithm \cite{may2020marc}, a widely recognized model for the trajectory user linking task. MARC utilizes one-hot encoded metadata, including departure day and hour, as well as geo-hashed trajectory information, to predict the probability that a given trajectory belongs to each user in the dataset. This evaluation is conducted exclusively on the Porto dataset, as it includes taxi identifiers, while the Chengdu dataset lacks user information, preventing its use in this evaluation. Moreover, the privacy evaluation is focused solely on seed synthetic trajectories generated by diffusion based models because the other baseline models generate trajectories based on their learned distribution, without leveraging metadata from the original trajectories for guidance. Randomly assigning other baseline-generated trajectories to a taxi ID does not accurately reflect the intended privacy level for this evaluation. 
\begin{figure}[t]

    \centering
    \includegraphics[width=\linewidth]{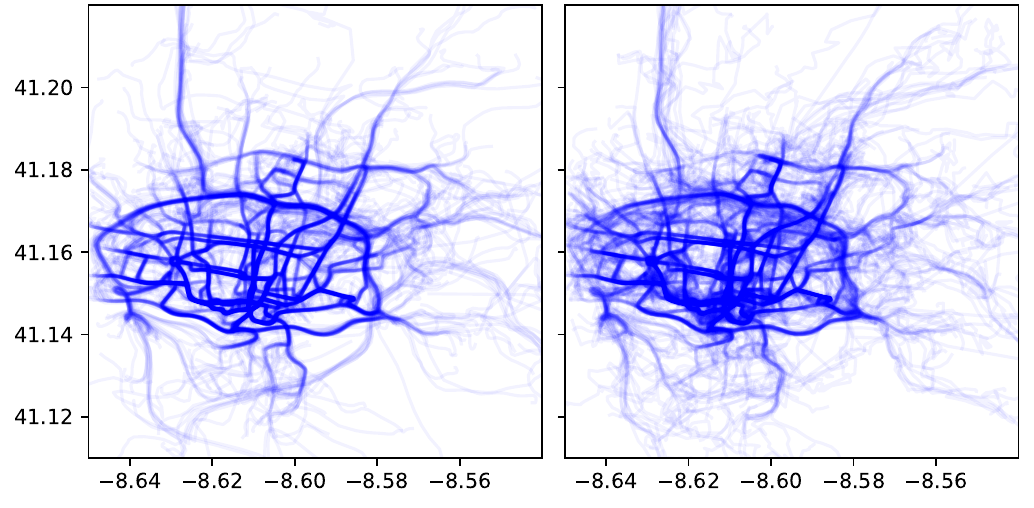}
    \caption{Left: Samples of Synthetic Trajectories by DiffTraj; Right: Samples of Synthetic Trajectories by \modelnamespace. As evidenced by the comparison, \modelnamespace preserves all essential mobility movement patterns while generating more diverse trajectories and hence offering more privacy protection.}
    \label{fig:porto-compare}
\vspace{-20pt}
\end{figure}

As shown in Table \ref{tab:privacy}, \modelnamespace consistently obtained a lower score across all metrics compared to DiffTraj. The key differentiator for \modelnamespace is its approach to noise sampling at two distinct levels of the trajectory generation process. By injecting noise at both the higher structural level and the finer granular level, \modelnamespace adds an additional layer of randomness and diversity to the synthetic trajectories. This layered noise application ensures that the synthetic trajectories deviate more significantly from the original data without compromising the realistic nature of the generated trajectories as evidenced in figure \ref{fig:porto-compare}. As \modelname's generated trajectories are a composition of segments independently sampled under the guidance of generated milestones, such formulation makes it harder for adversaries to reverse engineer the original metadata.  \modelnamespace strikes a balance between achieving a low distribution similarity and preserving privacy with high utility such that the synthetic trajectories remain useful in downstream applications.

\begin{table*}[t]
  \caption{Ablation Study on variants of \modelnamespace on Porto Dataset. "." stands for Error. The composition of \modelnamespace outperforms all other variants. HDE stands for Harversine Distance Error for taxi destination. For Anomalous Mobility Pattern Similarity, GH denotes the cosine similarity between anomalous mobility patterns in global hourly scale, WH denotes the cosine similarity between anomalous mobility patterns in weekly hourly scale.} 
  \label{tab:ablation}
  \centering

  \begin{tabular}{cccccccccccc}
    \toprule
     Model & Density . ($\downarrow$)& Distance . ($\downarrow$)& Trip . ($\downarrow$) & Pattern Score ($\uparrow$) & HDE($\downarrow$) & GH ($\uparrow$) & WH ($\uparrow$)\\
    \midrule
     \modelname& \textbf{0.0015}& \textbf{0.0002}&\textbf{0.0079}& \textbf{0.91}& \textbf{2.361}& \textbf{0.846} & \textbf{0.700}\\
     open angle extraction & 0.0073 & 0.0008 & 0.0178 & 0.79 & 2.388& 0.826 & 0.656 \\
    precise location guidance & 0.0039 & 0.0396 & 0.0804 & 0.78& 2.375 & 0.815 & 0.623\\
    global milestone & 0.0038 & 0.0372 & 0.0799 & 0.87 & 2.3966& 0.81 & 0.656\\
    \midrule
    Original Data &-&-&-&-& 2.319 & 1 & 1 \\
    \bottomrule
  \end{tabular}
  
\end{table*}

\section{Ablation study}

Table \ref{tab:ablation} reports four configuration of \modelnamespace evaluated on Porto Dataset: \textit{\modelnamespace} denotes the original configuration, \textit{open angle extraction} denotes the milestones are extracted based on the open angle detection algorithm instead of uniform sampling from the original trajectory, \textit{precise location guidance} denotes that the start and end location guidance incorporated as conditional information in both diffusion models are represented as precise location instead of being mapped to a discrete grid location, \textit{global milestone} denotes the configuration where the whole milestone trajectory $[m_1, m_2, ..., m_k]$ are given to generate segments instead of only using two milestones $[m_i,m_{i+1}]$ as start and end guidance to generate the $i$-th segment in a trajectory.  We note that all variants of \modelnamespace did not perform as well as \modelnamespace in both distribution similarity and downstream utility, validating the choice of design for \modelname. 

In addition, we further performed analysis on how the uniform sampling rate affects the performance of \modelnamespace on distribution similarity and downstream utility. We conducted analysis on the Chengdu dataset with uniform sampling rate 15, 35, and 55 as shown in table \ref{tab:sampling_rate}. We did not include 5 in the analysis because the Chengdu trajectories are densely sampled and extracting milestones every 5 timestamps provide virtually no level of abstraction to the original trajectories. In general, there is no apparent pattern that dictates an optimal rate universally. The analysis demonstrates that a sampling rate of 15 generally results in better performance across most metrics, suggesting that the lowest level of abstraction provides an optimal balance between data abstraction and fidelity.

\begin{table}[htbp]
  \caption{Analysis on the effect of uniform sampling rate 15, 35, and 55 on the Chengdu Dataset. HDE stands for harversine distance error and GH denotes the global hourly scale. The uniform sampling rate 5 is not considered in this case because Chengdu trajectories are sampled very densely and sampling every 5 points provide virtually no level of abstraction }
  \label{tab:sampling_rate}
  \centering
  \begin{tabular}{cccc}
    \toprule
    Metrics & \modelnamespace 15 & \modelnamespace 35 & \modelnamespace 55 \\
    \midrule
    Density Error & 0.0028 & 0.0337 & 0.0099 \\
    Distance Error & 0.0035 & 0.0423 & 0.0095 \\
    Trip Error & 0.0134 & 0.1698 & 0.0582 \\
    Patter Score & 0.81 & 0.6 & 0.71 \\
    HDE & 2.833 & 2.855 & 2.861 \\
    GH Cosine Similarity& 0.747 & 0.514 & 0.68 \\
  \bottomrule
\end{tabular}
\vspace{-20pt}
\end{table}

\section{Conclusion}

In this paper, we presented \modelname, a multi-resolution diffusion framework for synthetic human mobility trajectory generation. By modeling milestone-level global structure and segment-level local refinement, \modelnamespace 
 captures multi-scale spatial-temporal dependencies, enabling realistic long-range movement patterns while preserving fine-grained trajectory details.

Extensive experiments on large-scale real-world taxi datasets show that \modelnamespace achieves performance comparable to state-of-the-art diffusion baselines in global distribution similarity while consistently improving fidelity at finer spatial resolutions. The results highlight the importance of explicitly modeling local trajectory segments for high-quality synthetic data generation.

Beyond distributional metrics, the generated trajectories demonstrate strong downstream utility on tasks such as taxi destination prediction and anomalous mobility pattern identification, indicating the model’s ability to retain informative mobility behaviors. Under a seed-guided release setting, multi-level stochasticity further reduces linkage success compared to single-level diffusion approaches without sacrificing utility.

Overall, \modelnamespace provides an effective framework for generating realistic and useful synthetic mobility data while improving resistance to linkage attacks. This work underscores the potential of multi-resolution generative modeling for balancing realism, utility, and privacy in synthetic spatiotemporal data.

\section{Limitations and Ethical Considerations}

Despite the strong empirical performance of MR-Traj, several limitations remain. Although multi-level noise injection reduces trajectory linkage risk, synthetic data may still expose residual privacy information under stronger adversarial settings. Appropriate safeguards should therefore accompany any data release.

MR-Traj is trained on real-world taxi datasets from specific urban regions, which may introduce geographic and behavioral biases and limit generalizability to other cities or mobility contexts. In addition, metadata-conditioned generation may propagate biases present in the original data.

Finally, while synthetic trajectories can support downstream analysis, they should not replace real-world data in safety-critical applications such as policy-making or infrastructure planning. We encourage responsible use of synthetic mobility data and continued research on improving privacy protection and robustness.

\section{GenAI Disclosure}
Generative AI tools were used solely for language polishing and improving the clarity of presentation. They were not involved in the development of the methodology, data processing, experimental design, result analysis, or the derivation of scientific conclusions. All authors take full responsibility for the content of this paper.

\begin{acks}
This work is partially supported by the NSF Award 2425919, and NSF Award 2413417. The funding from these sources has been a cornerstone in enabling us to bring our project to fruition. We are also deeply grateful to the anonymous reviewers for their rigorous review process. Their detailed comments and constructive suggestions have significantly contributed to the improvement of this paper.
\end{acks}

\bibliographystyle{ACM-Reference-Format}
\bibliography{ref}

@inproceedings{zhu2023difftraj,
  title={DiffTraj: Generating GPS Trajectory with Diffusion Probabilistic Model},
  author={Zhu, Yuanshao and Ye, Yongchao and Zhang, Shiyao and Zhao, Xiangyu and Yu, James},
  booktitle={Thirty-seventh Conference on Neural Information Processing Systems},
  year={2023}
}

@article{song2003evaluating,
  title={Evaluating location predictors with extensive Wi-Fi mobility data},
  author={Song, Libo and Kotz, David and Jain, Ravi and He, Xiaoning},
  journal={ACM SIGMOBILE Mobile Computing and Communications Review},
  volume={7},
  number={4},
  pages={64--65},
  year={2003},
  publisher={ACM New York, NY, USA}
}

@article{yin2017generative,
  title={A generative model of urban activities from cellular data},
  author={Yin, Mogeng and Sheehan, Madeleine and Feygin, Sidney and Paiement, Jean-Fran{\c{c}}ois and Pozdnoukhov, Alexei},
  journal={IEEE Transactions on Intelligent Transportation Systems},
  volume={19},
  number={6},
  pages={1682--1696},
  year={2017},
  publisher={IEEE}
}

@article{hochreiter1997long,
  title={Long short-term memory},
  author={Hochreiter, Sepp and Schmidhuber, J{\"u}rgen},
  journal={Neural computation},
  volume={9},
  number={8},
  pages={1735--1780},
  year={1997},
  publisher={MIT press}
}

@inproceedings{nichol2021improved,
  title={Improved denoising diffusion probabilistic models},
  author={Nichol, Alexander Quinn and Dhariwal, Prafulla},
  booktitle={International Conference on Machine Learning},
  pages={8162--8171},
  year={2021},
  organization={PMLR}
}

@article{sun2016overview,
  title={An overview of moving object trajectory compression algorithms},
  author={Sun, Penghui and Xia, Shixiong and Yuan, Guan and Li, Daxing and others},
  journal={Mathematical Problems in Engineering},
  volume={2016},
  year={2016},
  publisher={Hindawi}
}

@inproceedings{lv2018t,
  title={T-CONV: A convolutional neural network for multi-scale taxi trajectory prediction},
  author={Lv, Jianming and Li, Qing and Sun, Qinghui and Wang, Xintong},
  booktitle={2018 IEEE international conference on big data and smart computing (bigcomp)},
  pages={82--89},
  year={2018},
  organization={IEEE}
}

@article{xu2019anomalous,
  title={Anomalous urban mobility pattern detection based on GPS trajectories and POI data},
  author={Xu, Zhenzhou and Cui, Ge and Zhong, Ming and Wang, Xin},
  journal={ISPRS International Journal of Geo-Information},
  volume={8},
  number={7},
  pages={308},
  year={2019},
  publisher={MDPI}
}

@inproceedings{pi2023hierarchical,
  title={Hierarchical generation of human-object interactions with diffusion probabilistic models},
  author={Pi, Huaijin and Peng, Sida and Yang, Minghui and Zhou, Xiaowei and Bao, Hujun},
  booktitle={Proceedings of the IEEE/CVF International Conference on Computer Vision},
  pages={15061--15073},
  year={2023}
}

@article{yang2024survey,
  title={A survey on diffusion models for time series and spatio-temporal data},
  author={Yang, Yiyuan and Jin, Ming and Wen, Haomin and Zhang, Chaoli and Liang, Yuxuan and Ma, Lintao and Wang, Yi and Liu, Chenghao and Yang, Bin and Xu, Zenglin and others},
  journal={ACM Computing Surveys},
  year={2024},
  publisher={ACM New York, NY}
}

@inproceedings{qiang2023coarse,
  title={Coarse-to-fine: a hierarchical diffusion model for molecule generation in 3d},
  author={Qiang, Bo and Song, Yuxuan and Xu, Minkai and Gong, Jingjing and Gao, Bowen and Zhou, Hao and Ma, Wei-Ying and Lan, Yanyan},
  booktitle={International Conference on Machine Learning},
  pages={28277--28299},
  year={2023},
  organization={PMLR}
}

@article{may2020marc,
  title={MARC: a robust method for multiple-aspect trajectory classification via space, time, and semantic embeddings},
  author={May Petry, Lucas and Leite Da Silva, Camila and Esuli, Andrea and Renso, Chiara and Bogorny, Vania},
  journal={International Journal of Geographical Information Science},
  volume={34},
  number={7},
  pages={1428--1450},
  year={2020},
  publisher={Taylor \& Francis}
}

@inproceedings{rombach2022high,
  title={High-resolution image synthesis with latent diffusion models},
  author={Rombach, Robin and Blattmann, Andreas and Lorenz, Dominik and Esser, Patrick and Ommer, Bj{\"o}rn},
  booktitle={Proceedings of the IEEE/CVF conference on computer vision and pattern recognition},
  pages={10684--10695},
  year={2022}
}

@article{zhu2024synmob,
  title={SynMob: Creating High-Fidelity Synthetic GPS Trajectory Dataset for Urban Mobility Analysis},
  author={Zhu, Yuanshao and Ye, Yongchao and Wu, Ying and Zhao, Xiangyu and Yu, James},
  journal={Advances in Neural Information Processing Systems},
  volume={36},
  year={2024}
}

@inproceedings{liu2018trajgans,
  title={trajGANs: Using generative adversarial networks for geo-privacy protection of trajectory data (Vision paper)},
  author={Liu, Xi and Chen, Hanzhou and Andris, Clio},
  booktitle={Location privacy and security workshop},
  pages={1--7},
  year={2018}
}

@article{rao2020lstm,
  title={LSTM-TrajGAN: A deep learning approach to trajectory privacy protection},
  author={Rao, Jinmeng and Gao, Song and Kang, Yuhao and Huang, Qunying},
  journal={arXiv preprint arXiv:2006.10521},
  year={2020}
}

@article{kang2020trag,
  title={TraG: A trajectory generation technique for simulating urban crowd mobility},
  author={Kang, Xu and Liu, Liang and Zhao, Dong and Ma, Huadong},
  journal={IEEE Transactions on Industrial Informatics},
  volume={17},
  number={2},
  pages={820--829},
  year={2020},
  publisher={IEEE}
}

@article{wang2021large,
  title={Large scale GPS trajectory generation using map based on two stage GAN},
  author={Wang, Xingrui and Liu, Xinyu and Lu, Ziteng and Yang, Hanfang},
  journal={Journal of Data Science},
  volume={19},
  number={1},
  pages={126--141},
  year={2021},
  publisher={中華資料採礦協會}
}

@article{choi2021trajgail,
  title={TrajGAIL: Generating urban vehicle trajectories using generative adversarial imitation learning},
  author={Choi, Seongjin and Kim, Jiwon and Yeo, Hwasoo},
  journal={Transportation Research Part C: Emerging Technologies},
  volume={128},
  pages={103091},
  year={2021},
  publisher={Elsevier}
}

@article{zhang2023dp,
  title={Dp-trajgan: A privacy-aware trajectory generation model with differential privacy},
  author={Zhang, Jing and Huang, Qihan and Huang, Yirui and Ding, Qian and Tsai, Pei-Wei},
  journal={Future Generation Computer Systems},
  volume={142},
  pages={25--40},
  year={2023},
  publisher={Elsevier}
}

@inproceedings{feng2020learning,
  title={Learning to simulate human mobility},
  author={Feng, Jie and Yang, Zeyu and Xu, Fengli and Yu, Haisu and Wang, Mudan and Li, Yong},
  booktitle={Proceedings of the 26th ACM SIGKDD international conference on knowledge discovery \& data mining},
  pages={3426--3433},
  year={2020}
}

@inproceedings{yu2017seqgan,
  title={Seqgan: Sequence generative adversarial nets with policy gradient},
  author={Yu, Lantao and Zhang, Weinan and Wang, Jun and Yu, Yong},
  booktitle={Proceedings of the AAAI conference on artificial intelligence},
  volume={31},
  number={1},
  year={2017}
}

@inproceedings{long2023practical,
  title={Practical synthetic human trajectories generation based on variational point processes},
  author={Long, Qingyue and Wang, Huandong and Li, Tong and Huang, Lisi and Wang, Kun and Wu, Qiong and Li, Guangyu and Liang, Yanping and Yu, Li and Li, Yong},
  booktitle={Proceedings of the 29th ACM SIGKDD Conference on Knowledge Discovery and Data Mining},
  pages={4561--4571},
  year={2023}
}

@inproceedings{ronneberger2015u,
  title={U-net: Convolutional networks for biomedical image segmentation},
  author={Ronneberger, Olaf and Fischer, Philipp and Brox, Thomas},
  booktitle={Medical Image Computing and Computer-Assisted Intervention--MICCAI 2015: 18th International Conference, Munich, Germany, October 5-9, 2015, Proceedings, Part III 18},
  pages={234--241},
  year={2015},
  organization={Springer}
}

@misc{pkdd-15-predict-taxi-service-trajectory-i,
    author = {Meghan O'Connell, moreiraMatias, Wendy Kan},
    title = {ECML/PKDD 15: Taxi Trajectory Prediction (I)},
    publisher = {Kaggle},
    year = {2015},
    url = {https://kaggle.com/competitions/pkdd-15-predict-taxi-service-trajectory-i}
}

@book{villani2009optimal,
  title={Optimal transport: old and new},
  author={Villani, C{\'e}dric and others},
  volume={338},
  year={2009},
  publisher={Springer}
}

@book{devroye2013probabilistic,
  title={A probabilistic theory of pattern recognition},
  author={Devroye, Luc and Gy{\"o}rfi, L{\'a}szl{\'o} and Lugosi, G{\'a}bor},
  volume={31},
  year={2013},
  publisher={Springer Science \& Business Media}
}

@book{boyd2004convex,
  title={Convex optimization},
  author={Boyd, Stephen and Vandenberghe, Lieven},
  year={2004},
  publisher={Cambridge university press}
}

@article{schneider2013unravelling,
  title={Unravelling daily human mobility motifs},
  author={Schneider, Christian M and Belik, Vitaly and Couronn{\'e}, Thomas and Smoreda, Zbigniew and Gonz{\'a}lez, Marta C},
  journal={Journal of The Royal Society Interface},
  volume={10},
  number={84},
  pages={20130246},
  year={2013},
  publisher={The Royal Society}
}

@article{song2020denoising,
  title={Denoising diffusion implicit models},
  author={Song, Jiaming and Meng, Chenlin and Ermon, Stefano},
  journal={arXiv preprint arXiv:2010.02502},
  year={2020}
}

@inproceedings{bindschaedler2016synthesizing,
  title={Synthesizing plausible privacy-preserving location traces},
  author={Bindschaedler, Vincent and Shokri, Reza},
  booktitle={2016 IEEE Symposium on Security and Privacy (SP)},
  pages={546--563},
  year={2016},
  organization={IEEE}
}

@article{doersch2016tutorial,
  title={Tutorial on variational autoencoders},
  author={Doersch, Carl},
  journal={arXiv preprint arXiv:1606.05908},
  year={2016}
}

@article{creswell2018generative,
  title={Generative adversarial networks: An overview},
  author={Creswell, Antonia and White, Tom and Dumoulin, Vincent and Arulkumaran, Kai and Sengupta, Biswa and Bharath, Anil A},
  journal={IEEE signal processing magazine},
  volume={35},
  number={1},
  pages={53--65},
  year={2018},
  publisher={IEEE}
}

@article{sherstinsky2020fundamentals,
  title={Fundamentals of recurrent neural network (RNN) and long short-term memory (LSTM) network},
  author={Sherstinsky, Alex},
  journal={Physica D: Nonlinear Phenomena},
  volume={404},
  pages={132306},
  year={2020},
  publisher={Elsevier}
}

@article{arab2008hierarchical,
  title={Hierarchical Spatial Models.},
  author={Arab, Ali and Hooten, Mevin B and Wikle, Christopher K},
  journal={Encyclopedia of GIS},
  volume={14},
  number={1},
  pages={425--431},
  year={2008}
}

@article{li2020trajectory,
  title={Trajectory data-based traffic flow studies: A revisit},
  author={Li, Li and Jiang, Rui and He, Zhengbing and Chen, Xiqun Michael and Zhou, Xuesong},
  journal={Transportation Research Part C: Emerging Technologies},
  volume={114},
  pages={225--240},
  year={2020},
  publisher={Elsevier}
}

@article{kraemer2020effect,
  title={The effect of human mobility and control measures on the COVID-19 epidemic in China},
  author={Kraemer, Moritz UG and Yang, Chia-Hung and Gutierrez, Bernardo and Wu, Chieh-Hsi and Klein, Brennan and Pigott, David M and Open COVID-19 Data Working Group† and Du Plessis, Louis and Faria, Nuno R and Li, Ruoran and others},
  journal={Science},
  volume={368},
  number={6490},
  pages={493--497},
  year={2020},
  publisher={American Association for the Advancement of Science}
}

@article{guo2025leveraging,
  title={Leveraging the Spatial Hierarchy: Coarse-to-fine Trajectory Generation via Cascaded Hybrid Diffusion},
  author={Guo, Baoshen and Hong, Zhiqing and Li, Junyi and Wang, Shenhao and Zhao, Jinhua},
  journal={arXiv preprint arXiv:2507.13366},
  year={2025}
}

@article{luo2025traveller,
  title={Traveller: Travel-pattern aware trajectory generation via autoregressive diffusion models},
  author={Luo, Yuxiao and Zhang, Songming and Liu, Kang and Xu, Yang and Yin, Ling},
  journal={Information Fusion},
  pages={103766},
  year={2025},
  publisher={Elsevier}
}

@inproceedings{liu2025multi,
  title={Multi-scale diffusion transformer for jointly simulating user mobility and mobile traffic pattern},
  author={Liu, Ziyi and Long, Qingyue and Wang, Huandong and Li, Yong},
  booktitle={Proceedings of the 33rd ACM International Conference on Advances in Geographic Information Systems},
  pages={522--525},
  year={2025}
}

@article{dirmeier2024synthetic,
  title={Synthetic location trajectory generation using categorical diffusion models},
  author={Dirmeier, Simon and Hong, Ye and Perez-Cruz, Fernando},
  journal={arXiv preprint arXiv:2402.12242},
  year={2024}
}

@article{song2024controllable,
  title={Controllable human trajectory generation using profile-guided latent diffusion},
  author={Song, Yiwen and Ding, Jingtao and Yuan, Jian and Liao, Qingmin and Li, Yong},
  journal={ACM Transactions on Knowledge Discovery from Data},
  volume={19},
  number={1},
  pages={1--25},
  year={2024},
  publisher={ACM New York, NY}
}

@article{kapp2023generative,
  title={Generative models for synthetic urban mobility data: A systematic literature review},
  author={Kapp, Alexandra and Hansmeyer, Julia and Mihaljevi{\'c}, Helena},
  journal={ACM Computing Surveys},
  volume={56},
  number={4},
  pages={1--37},
  year={2023},
  publisher={ACM New York, NY}
}

@article{graser2025mobilitydl,
  title={MobilityDL: a review of deep learning from trajectory data},
  author={Graser, Anita and Jalali, Anahid and Lampert, Jasmin and Wei{\ss}enfeld, Axel and Janowicz, Krzysztof},
  journal={GeoInformatica},
  volume={29},
  number={1},
  pages={115--147},
  year={2025},
  publisher={Springer}
}

@article{wu2024pretrained,
  title={Pretrained mobility transformer: A foundation model for human mobility},
  author={Wu, Xinhua and He, Haoyu and Wang, Yanchao and Wang, Qi},
  journal={arXiv preprint arXiv:2406.02578},
  year={2024}
}

@article{zhu2024unitraj,
  title={Unitraj: Learning a universal trajectory foundation model from billion-scale worldwide traces},
  author={Zhu, Yuanshao and Yu, James Jianqiao and Zhao, Xiangyu and Zhou, Xun and Han, Liang and Wei, Xuetao and Liang, Yuxuan},
  journal={arXiv preprint arXiv:2411.03859},
  year={2024}
}

@article{garg2025gps,
  title={GPS-MTM: Capturing Pattern of Normalcy in GPS-Trajectories with self-supervised learning},
  author={Garg, Umang and Zhang, Bowen and Subrahmanya, Anantajit and Gudavalli, Chandrakanth and Manjunath, BS},
  journal={arXiv preprint arXiv:2509.24031},
  year={2025}
}

@article{cao2025timedit,
  title={TimeDiT: General-purpose diffusion transformers for time series foundation model},
  author={Cao, Defu and Ye, Wen and Zhang, Yizhou and Liu, Yan},
  journal={arXiv preprint arXiv:2409.02322},
  year={2024}
}

@INPROCEEDINGS{9561461,
  author={Cao, Defu and Li, Jiachen and Ma, Hengbo and Tomizuka, Masayoshi},
  booktitle={2021 IEEE International Conference on Robotics and Automation (ICRA)}, 
  title={Spectral Temporal Graph Neural Network for Trajectory Prediction}, 
  year={2021},
  volume={},
  number={},
  pages={1839-1845},
  doi={10.1109/ICRA48506.2021.9561461}}

@article{
ye2024domain,
title={{TS}-Reasoner: Domain-Oriented Time Series Inference Agents for Reasoning and Automated Analysis},
author={Wen Ye and Wei Yang and Defu Cao and Yizhou Zhang and Lumingyuan Tang and Jie Cai and Yan Liu},
journal={Transactions on Machine Learning Research},
issn={2835-8856},
year={2026},
url={https://openreview.net/forum?id=yhy7Vigjcf},
note={}
}

@article{cao2021spectral,
  title={Spectral temporal graph neural network for multivariate time-series forecasting},
  author={Cao, Defu and Wang, Yujing and Duan, Juanyong and Zhang, Ce and Zhu, Xia and Huang, Conguri and Tong, Yunhai and Xu, Bixiong and Bai, Jing and Tong, Jie and others},
  journal={Advances in neural information processing systems},
  year={2021}
}

@article{cao2025conversational,
  title={Conversational Time Series Foundation Models: Towards Explainable and Effective Forecasting},
  author={Cao, Defu and Gee, Michael and Liu, Jinbo and Wang, Hengxuan and Yang, Wei and Wang, Rui and Liu, Yan},
  journal={arXiv preprint arXiv:2512.16022},
  year={2025}
}

@inproceedings{
cao2026pinfdit,
title={{PINFD}iT: Energy-Based Physics-Informed Diffusion Transformers for General-purpose Time Series Tasks},
author={Defu Cao and Wen Ye and Yizhou Zhang and Sam Griesemer and Yan Liu},
booktitle={The Fourteenth International Conference on Learning Representations},
year={2026},
url={https://openreview.net/forum?id=EphTlUJ4XN}
}

@inproceedings{
cao2023tempo,
title={{TEMPO}: Prompt-based Generative Pre-trained Transformer for Time Series Forecasting},
author={Defu Cao and Furong Jia and Sercan O Arik and Tomas Pfister and Yixiang Zheng and Wen Ye and Yan Liu},
booktitle={The Twelfth International Conference on Learning Representations},
year={2024},
url={https://openreview.net/forum?id=YH5w12OUuU}
}

@inproceedings{
griesemer2024active,
title={Active Sequential Posterior Estimation for Sample-Efficient Simulation-Based Inference},
author={Sam Griesemer and Defu Cao and Zijun Cui and Carolina Osorio and Yan Liu},
booktitle={The Thirty-eighth Annual Conference on Neural Information Processing Systems},
year={2024},
url={https://openreview.net/forum?id=fkuseU0nJs}
}

@article{zhang2022counterfactual,
  title={Counterfactual neural temporal point process for estimating causal influence of misinformation on social media},
  author={Zhang, Yizhou and Cao, Defu and Liu, Yan},
  journal={Advances in Neural Information Processing Systems},
  volume={35},
  pages={10643--10655},
  year={2022}
}

@inproceedings{niu2022mu2rest,
  title={Mu2ReST: Multi-resolution Recursive Spatio-Temporal Transformer for Long-Term Prediction},
  author={Niu, Hao and Meng, Chuizheng and Cao, Defu and Habault, Guillaume and Legaspi, Roberto and Wada, Shinya and Ono, Chihiro and Liu, Yan},
  booktitle={Pacific-Asia Conference on Knowledge Discovery and Data Mining},
  pages={68--80},
  year={2022}
}

@article{meng2025physics,
  title={When physics meets machine learning: A survey of physics-informed machine learning},
  author={Meng, Chuizheng and Griesemer, Sam and Cao, Defu and Seo, Sungyong and Liu, Yan},
  journal={Machine Learning for Computational Science and Engineering},
  volume={1},
  number={1},
  pages={20},
  year={2025},
  publisher={Springer}
}

@article{cao2022dslob,
  title={DSLOB: a synthetic limit order book dataset for benchmarking forecasting algorithms under distributional shift},
  author={Cao, Defu and El-Laham, Yousef and Trinh, Loc and Vyetrenko, Svitlana and Liu, Yan},
  journal={arXiv preprint arXiv:2211.11513},
  year={2022}
}

\appendix

\section{Dataset Information}
\label{dataset_section}
The two large-scale taxi trajectory datasets used in this work are the Chengdu dataset and the Porto dataset. The Porto taxi trajectory dataset was released as part of the ECML/PKDD 2015 challenge on taxi destination prediction. It was collected in the city of Porto over the course of one year and contains over one million individual trajectories. GPS points are recorded at a sampling interval of 15 seconds. 

\begin{figure}
    \centering
    \includegraphics[width=\linewidth]{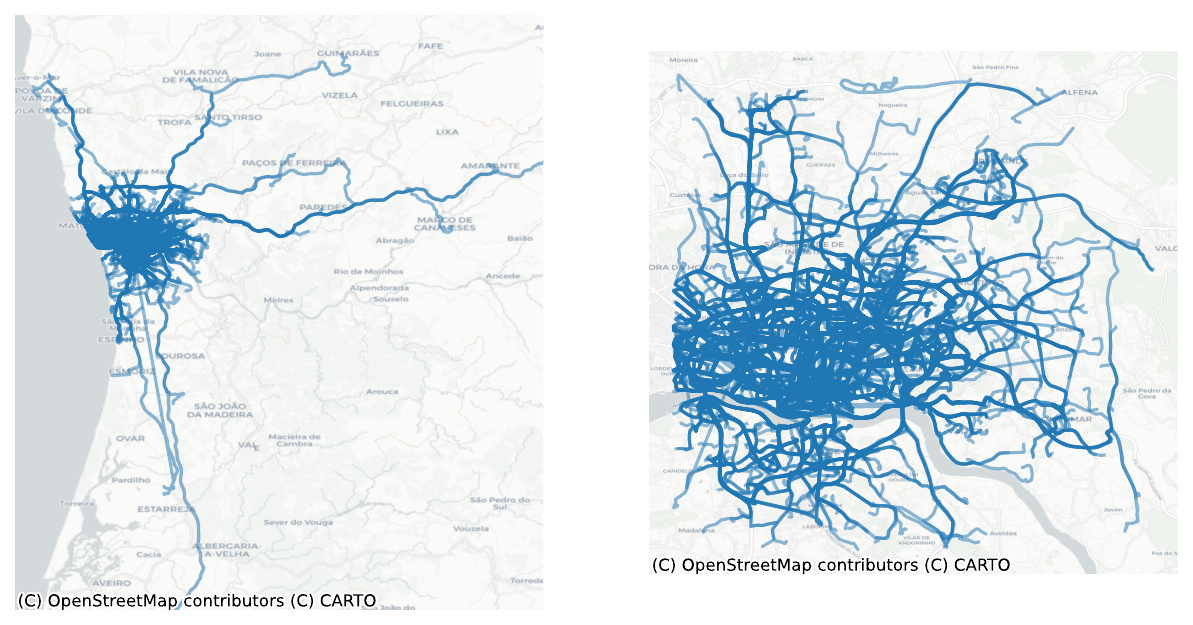}
    \caption{Left: samples from original trajectories; Right: filtered samples within the kernel region of the city of Porto.}
    \label{fig:kernelregion}
\end{figure}

We filtered out trajectories with fewer than 10 points and focused on trajectories within the central area of Porto, commonly referred to as the kernel region of a city~\cite{kang2020trag}. This region exhibits higher and more uniform trajectory density, providing a representative subset for analysis. An illustration of the kernel region is shown in Figure~\ref{fig:kernelregion}. 

The Chengdu dataset is already confined to a high-density urban region, and no additional spatial filtering is applied~\cite{zhu2024synmob}. Table~\ref{tab:dataset} summarizes the key statistics of the processed datasets, including the number of trajectories, trajectory lengths, and spatial coverage.

\begin{table}
  \caption{Dataset Information.}
  \label{tab:dataset}
  \centering
  \scalebox{0.8}{
  \begin{tabular}{ccc}
    \toprule
    Meta-Information & Filtered Porto & Chengdu\\
    \midrule
    \# of records & 112991 & 1000000\\
    \# of taxi id & 441 & -\\
    min/max/mean length of records & 10/2198/44.07 & 180/6890/292.42 \\
    min/max latitude & 41.1 / 41.25 & 30.65 / 30.74 \\
    min/max longitude & -8.65 / -8.5 & 104.03 / 104.13 \\
    sampling frequency & 15s & 15s \\
    \midrule
    Local Case Study & Ribeira, Porto & Taikoo Li, Chengdu\\
    min/max latitude & 41.140344 / 41.143130 & 30.651997 / 30.661530 \\
    min/max longitude & -8.615496 / -8.606797 & 104.075045 / 104.089312 \\
    \# of segments & 99601 & 681038 \\
  \bottomrule
\end{tabular}}
\end{table}

\section{Evaluation Metrics}
\label{metrics}
This section formally defines the evaluation metrics used in the main experiments.

\begin{itemize}
    \item \textbf{Density Error} measures the discrepancy between the spatial distributions of synthetic and real trajectories over a discretized grid. It is computed using the Jensen--Shannon divergence (JSD) between the two distributions:
    \begin{equation}
        \textit{Density Error} = \mathcal{JSD}(\mathcal{D}(S_{syn}), \mathcal{D}(S_{real})).
    \end{equation}

    \item \textbf{Distance Error} evaluates the difference in total travel distance distributions. The total distance of a trajectory is computed as the sum of distances between consecutive GPS points. The divergence between the distributions of synthetic and real trajectories is measured using JSD:
    \begin{equation}
        \textit{Distance Error} = \mathcal{JSD}(Dis(S_{syn}), Dis(S_{real})).
    \end{equation}

    \item \textbf{Pattern Score} quantifies the similarity between frequent movement patterns extracted from synthetic and real trajectories. A pattern is defined as a frequently occurring sequence of discretized grid cells. The similarity is computed using the F1 score:
    \begin{equation}
        \textit{Pattern F1} = 
        2 \times \frac{Precision(P_{real}, P_{syn}) \times Recall(P_{real}, P_{syn})}
        {Precision(P_{real}, P_{syn}) + Recall(P_{real}, P_{syn})}.
    \end{equation}
\end{itemize}

\section{Details of Downstream Tasks}

\paragraph{Taxi Destination Prediction}
\label{tconv_details}
T-CONV models trajectories as two-dimensional spatial paths and applies multi-scale convolutional operations to extract movement patterns. Trajectory destinations are first clustered into arrival hotspots, and the predicted destination is computed as a weighted combination of the corresponding cluster centers. We adopt the locally enhanced variant of T-CONV, which performs higher-resolution convolution near the start and end of trajectories.

The prediction error is measured using the Haversine distance, which computes the shortest distance between two points on the Earth’s surface:
\begin{equation}
  D(d_{real}, d_{pred}) = 2r \cdot \arctan\left(\sqrt{\frac{\alpha}{1-\alpha}}\right),
\end{equation}
where $d_{real}$ and $d_{pred}$ denote the ground-truth and predicted destinations, respectively, and $r$ is the Earth’s radius.

\paragraph{Anomalous Mobility Pattern Identification}
\label{ano_details}
This task aims to identify anomalous mobility patterns from a collection of trajectories by analyzing time-dependent spatial movement distributions. Mobility vectors are constructed for each time interval over a spatial grid:
\begin{equation}
    mv_i =
    \left(
    \begin{array}{c}
    f(t_i,g_1)\\
    f(t_i,g_2) \\
    \vdots \\
    f(t_i,g_k)
    \end{array}
    \right)
    \in \mathbb{R}^{k \times 1},
\end{equation}
where $k$ is the total number of grid cells and $f(t_i,g_j)$ counts trips departing at time $t_i$ with destinations in grid $g_j$. Each mobility vector is normalized by the total number of trips.

Anomalous mobility vectors are identified following the procedure in~\cite{xu2019anomalous}: (1) constructing mobility vectors at the specified temporal scale, (2) clustering vectors using a mean-shift algorithm with a flat kernel, (3) computing an anomaly threshold based on cluster size statistics, and (4) labeling vectors in clusters smaller than the threshold as anomalous.

\section{Hyperparameter Setting}
Hyperparameter configurations for training and inference are summarized in Table~\ref{tab:hyperparameter}. All diffusion models are trained on an 80GB A100 GPU, while downstream task models are trained on a 10GB RTX 2080 GPU.

\begin{table}
    \caption{Hyperparameter Details for Experiment Settings}
    \label{tab:hyperparameter}
    \centering
    \scalebox{0.75}{
    \begin{tabular}{ccc}
    \toprule
         Hyperparameter & Milestone Level Model & Segment Level Model \\
         \midrule
         Epoch & 200 & 100 \\
         Batch Size & 1024 & 1024 \\
         Noise Schedule & Cosine & Cosine \\
         Input Length (Porto) & 11 & 7 \\
         Input Length (Chengdu) & 23 & 23 \\
         Sampling Rate for Milestone (Porto) & 5 & - \\
         Sampling Rate for Milestone (Chengdu) & 15 & - \\
         Diffusion Timesteps & 500 & 500 \\
    \bottomrule
    \end{tabular}}
\end{table}

\section{Theoretic Analysis}
\label{theo}
The main application of synthetic trajectory generation is to train machine learning models for downstream tasks such as destination prediction, mileage prediction and anomalous mobility pattern recognition. These downstream tasks usually follow a paradigm to predict some properties (e.g., destination, mileage and mobility pattern) of a trip, denoted as $g(S_u,S_o)$, given only a fraction of the trip's full trajectory. We denote this observable fraction of full trajectory as $\mathcal{S}_o$. The rest unobservable fractions of the trajectory are denoted as $\mathcal{S}_u$. 

For downstream tasks, we usually use complete trajectories generated by the synthetic trajectory generator $P'(S)$ as training set to train the machine learning predictor $f(\cdot)$. 
The utility of this predictor on downstream tasks should be evaluated by its square empirical risk $\mathcal{R}$ on a trajectory distribution, which is defined as:
\begin{equation}
    \mathcal{R}(f(\cdot))=\mathbb{E}[f(S_o)-g(S_u,S_o)]^2=\mathbb{E}_{S_o}[f(S_o)-\mathbb{E}_{S_u}g(S_u,S_o)]^2+C
\end{equation}
where $C$ is a constant that is only relevant to $P(S_u,S_o)$. Thus, the optimal predictor we can acquire on the synthetic dataset is $f(S_o)=\mathbb{E}_{S_u\sim P'(S_u|S_o)}g(S_u,S_o)$.

This optimal predictor of synthetic dataset, when being applied to real dataset whose distribution is $P(S)$, its empirical risk will be:

\begin{equation}
\begin{aligned}
    \mathcal{R}\quad=\quad&\mathbb{E}_{S_o\sim P(S_o)}[\mathbb{E}_{S_u\sim P'(S_u|S_o)}g(S_u,S_o)\\
    &-\mathbb{E}_{S_u\sim P(S_u|S_o)}g(S_u,S_o)]^2+C\\
\end{aligned}
\end{equation}

Suppose that $g(S_u,S_o)$ is L-lipsitz, then $\frac{g(S_u,S_o)}{L}$ is 1-lipsitz, and we have:

\begin{equation}
\begin{aligned}
    \mathcal{R}\quad=\quad&L\mathbb{E}_{S_o\sim P(S_o)}[\mathbb{E}_{S_u\sim P'(S_u|S_o)}\frac{g(S_u,S_o)}{L}\\
    &-\mathbb{E}_{S_u\sim P(S_u|S_o)}\frac{g(S_u,S_o)}{L}]^2
    +C\\
    \leq\quad& L\mathbb{E}_{S_o\sim P(S_o)}[\sup_{h \text{ is 1-lipsitz}}\mathbb{E}_{S_u\sim P'(S_u|S_o)}h(S_u,S_o)\\
    &-\mathbb{E}_{S_u\sim P(S_u|S_o)}h(S_u,S_o)]^2+C\\
     = \quad&L\mathbb{E}_{S_o\sim P(S_o)}[W_1(P(S_u|S_o),P'(S_u|S_o))]^2+C\\
\end{aligned}
\end{equation}
where $W_1$ is Wasserstein distance of 1-norm. The above  For Wasserstein distance, we have following upper bound:
\begin{equation}
    W_1(P(S_u|S_o),P'(S_u|S_o))\leq D_{S_u}||P(S_u|S_o)-P'(S_u|S_o)||_{TV}
\end{equation}
where $D$ is the diameter of the support set of $S_u$ and $||P(S_u|S_o)-P'(S_u|S_o)||_{TV}$ is the Total Variance Distance between distribution $P(S_u|S_o)$ and $P'(S_u|S_o)$:
\begin{equation}
    ||P(S_u|S_o)-P'(S_u|S_o)||_{TV} = \int_{S_u} |p(S_u|S_o)-p'(S_u|S_o)| dS_u
\end{equation}

Thus, when Total Variance Distance between $P(S_u,S_o)$ and $P'(S_u,S_o)$ is less than 1, i.e., the two distributions are close enough, we have:
\begin{equation}
\begin{aligned}
    \mathcal{R}&\leq L\mathbb{E}_{S_o\sim P(S_o)}[W_1(P(S_u|S_o),P'(S_u|S_o))]^2+C\\
    &\leq LD^2\mathbb{E}_{S_o\sim P(S_o)}||P(S_u|S_o)-P'(S_u|S_o)||_{TV}+C
\end{aligned}
\end{equation}

For Total Variance Distance we have:
\begin{equation}
\begin{aligned}
    \mathbb{E}_{S_o\sim P(S_o)}TVD &= \mathbb{E}_{S_o\sim P(S_o)}||P(S_u|S_o)-P'(S_u|S_o)||_{TV} \\
    &= \int_{S_o} p(S_o)dS_o\int_{S_u}|p(S_u|S_o)-p'(S_u|S_o)| dS_u\\
    &=\int_{S_u,S_o} |p(S_u,S_o)-p'(S_u,S_o)| dS_udS_o\\
    &=||P(S_u,S_o)-P'(S_u,S_o)||_{TV}\\
\end{aligned}
\end{equation}

Meanwhile, Total Variance Distance satisfy triangle inequality. Thus, we have:
\begin{equation}
\begin{aligned}
    \mathcal{R} \quad\leq\quad& LD^2||P(S_u,S_o)-P'(S_u|S_o)P(S_o)||_{TV}+C \\
   \quad \leq\quad &LD^2||P(S_u,S_o)-P'(S_u,S_o)||_{TV}\\
    &+LD^2||P'(S_u,S_o)-P'(S_u|S_o)P(S_o)||_{TV}+C \\
    \quad=\quad& LD^2||P(S_u,S_o)-P'(S_u,S_o)||_{TV}\\
    &+LD^2||P'(S_u|S_o)P'(S_o)-P'(S_u|S_o)P(S_o)||_{TV}+C\\
    \quad=\quad& LD^2||P(S_u,S_o)-P'(S_u,S_o)||_{TV}+\\
    &LD^2||P'(S_o)-P(S_o)||_{TV}+C\\
\end{aligned}
\end{equation}
The above upper bound shows that we should jointly minimize $||P(S_u,S_o)-P'(S_u,S_o)||_{TV}$ and $||P'(S_o)-P(S_o)||_{TV}$.

\section{Details on Diffusion Model Preliminaries}

In recent years, denoising diffusion probabilistic model has gained tremendous success on generative tasks, particularly in image generation\cite{rombach2022high}. The allure of DDPMs lies in their capacity to model complex, high-dimensional distributions, making them an excellent choice for generating synthetic trajectories. The key idea behind diffusion is to gradually add noise to input over a series of timesteps and let the model learn the reverse process of denoising noisy input back to the realistic space. 
The forward process is defined as a Markov chain where Gaussian noise is gradually added to the original input $x_0$ and the amount of noise added at each timestep is controlled by the noise schedule $\beta_t$.
\begin{equation}
    q(x_{1:T} | x_0) = \prod_{t=1}^{T} q(x_t | x_{t-1})
\end{equation}
\begin{equation}
    q(x_t | x_{t-1}) = \mathcal{N}\left(x_t; \sqrt{1 - \beta_t} x_{t-1}, \beta_t I \right)
\end{equation}
In practice, the closed form sampling of the noised version of original input at timestep $t$ is used instead of adding noise step by step for efficiency purpose. 
\begin{equation}
    x_t = \sqrt{\overline{\alpha}_t} \cdot x_0 + \sqrt{1 - \overline{\alpha}_t} \cdot \epsilon , \quad \epsilon \sim \mathcal{N}(0, I)
\end{equation}
where $\overline{\alpha}_t = \prod_{i=1}^{t}1-\beta_t$. This closed form sampling allows for more efficient forward process. 
The backward process aims to learn the reverse of the forward process, learning to systematically denoise the noisy inputs. A neural network parameterized by $\theta$ is used to learn the mean and variance of the noise to remove from the noisy data at each timestep. The denoising process is described by:
\begin{equation}
    q(x_{t-1} | x_t) = \mathcal{N}(x_{t-1}; \mu(x_t, t), \Sigma(x_t, t))
\end{equation}
\begin{equation}
    p_{\theta}(x_{t-1}|x_t) = \mathcal{N}(x_{t-1}; \mu_{\theta}(x_t, t), \Sigma_{\theta}(x_t,t))
\end{equation}
where $p_{\theta}(x_{t-1}|x_t)$ is learned by the neural network to approximate the ground truth conditional probability $q(x_{t-1} | x_t)$.

\end{document}